\documentclass[runningheads]{llncs}
\usepackage{eccv}

\usepackage{eccvabbrv}

\usepackage[T1]{fontenc}
\usepackage{booktabs}
\usepackage{graphicx}
\usepackage{amsmath}
\usepackage{amssymb}
\usepackage{wrapfig}

\usepackage{hyperref}

\begin{document}
\title{DARES: Depth Anything in Robotic Endoscopic Surgery with Self-supervised Vector-LoRA of the Foundation Model}
%
%
\author{Mona Sheikh Zeinoddin \inst{1} \and
Chiara Lena\inst{3} \and
Jiongqi Qu\inst{2} \and
Luca Carlini \inst{3} \and
Mattia Magro\inst{3} \and
Seunghoi Kim \inst{2} \and
Elena De Momi \inst{3} \and
Sophia Bano \inst{1} \and
Matthew Grech-Sollars \inst{2} \and
Evangelos Mazomenos \inst{1} \and
Daniel C. Alexander \inst{2} \and
Danail Stoyanov \inst{1} \and
Matthew J. Clarkson \inst{1} \and
Mobarakol Islam \inst{1} }

\authorrunning{M. Zeinoddin et al.}
%
\institute{Wellcome/EPSRC Centre for Interventional and Surgical Sciences (WEISS), University College London, UK 
\and
Centre for Medical Image Computing, University College London, UK
\and
Dept. of Electronics, Information and Bioengineering, Politecnico di Milano, Milano, Italy}
\maketitle    
\begin{abstract}
Robotic-assisted surgery (RAS) relies on accurate depth estimation for 3D reconstruction and visualization. While foundation models like Depth Anything Models (DAM) show promise, directly applying them to surgery often yields suboptimal results. Fully fine-tuning on limited surgical data can cause overfitting and catastrophic forgetting, compromising model robustness and generalization. Although Low-Rank Adaptation (LoRA) addresses some adaptation issues, its uniform parameter distribution neglects the inherent feature hierarchy, where earlier layers, learning more general features, require more parameters than later ones. To tackle this issue, we introduce Depth Anything in Robotic Endoscopic Surgery (DARES), a novel approach that employs a new adaptation technique, Vector Low-Rank Adaptation (Vector-LoRA) on the DAM V2 to perform self-supervised monocular depth estimation in RAS scenes. To enhance learning efficiency, we introduce Vector-LoRA by integrating more parameters in earlier layers and gradually decreasing parameters in later layers. We also design a reprojection loss based on the multi-scale SSIM error to enhance depth perception by better tailoring the foundation model to the specific requirements of the surgical environment. The proposed method is validated on the SCARED dataset and demonstrates superior performance over recent state-of-the-art self-supervised monocular depth estimation techniques, achieving an improvement of 13.3\% in the absolute relative error metric. The code and pre-trained weights are available at~\url{https://github.com/mobarakol/DARES}.

\end{abstract}
\section{Introduction} \label{sec:intro}
Robot-assisted surgery (RAS) has been widely adopted in clinical practice to enhance operational precision and reduce physical discomfort \cite{privitera2022image}. In these procedures, depth estimation is essential for enabling high-definition visualisation \cite{wei2024absolute}, decision-making \cite{chadebecq2023artificial}, surgical navigation, and it enhances surgical outcomes by improving instrument insertion while reducing complications\cite{wang2023microsurgery}. Additionally, depth information is crucial for reconstructing reliable 3D models from 2D images, which aids in gaining deeper anatomical understanding, performing surgical planning \cite{yang20243d}, and serves as a fundamental step towards the use of augmented reality \cite{Fu2023}. 
Nevertheless, obtaining reliable depth information in endoscopic environments via traditional techniques such as Simultaneous Localisation and Mapping (SLAM) and Structure from Motion (SfM) remains a significant challenge, due to the limited field of view of the camera, low-light conditions, the presence of artifacts, textureless areas, and frequent occlusions \cite{CHEN2018}. 
Deep learning is a powerful tool to tackle these challenges and increase the accuracy and reliability of monocular depth estimation and 3D reconstruction algorithms. However, it requires extensive training with vast amounts of data, often unavailable in clinical practice. To tackle this challenge, a self-supervised learning (SSL) approach is introduced in \cite{shao2022self}, which extracts
robust depth and ego-motion from monocular endoscopic videos. Also, a new loss function is presented in this pipeline to deal with brightness variations typical of surgical scenes \cite{shao2022self}. 
However, the simple structure of this depth estimation architecture does not perfectly suit the complexities of RAS environments as will be later discussed.

Foundation models such as the Depth Anything Model (DAM) V1,V2\cite{yang2024depth,yang2024depthv2} and Surgical-DINO\cite{beilei2024surgical} 
have been pivotal in advancing depth estimation state-of-the-art (SOTA) methods. 
However, these models are not optimized for SSL. In addition, their training is excessively time-consuming and creates the risk of catastrophic forgetting for their learned knowledge \cite{luo2023empirical, wu2023medical}. Although parameter-efficient fine-tuning techniques, specifically Low-Rank Adaptation (LoRA) \cite{hu2022lora} have been introduced to solve these issues and adapt foundation models to domain-specific tasks, their uniform parameter distribution does not account for the feature hierarchy or gradient flow dynamics in deep networks. Earlier layers in these networks learn general features that require more parameters over later layers \cite{chai2021deep}. To address this issue, we introduce Vector-LoRA which allocates a unique rank to each layer, allowing a higher number of parameters in earlier layers of these networks. In addition, we design our SSL training scheme following a multi-scale SSIM based reprojection Loss, to better account for RAS scenes requirements.

Overall, we propose Depth Anything in Robotic Endoscopic Surgery (DARES), integrating Vector-LoRA into DAM V2 for monocular depth estimation in RAS scenes, and designing a self-supervised training scheme following a multi-scale SSIM based reprojection loss. The main contributions and findings of this work are: 
\begin{enumerate}

\item Introducing one of the very first works that adapts the full architecture of the DAM V2 to the surgical domain in an SSL manner to improve depth estimation without extensive labeled data.
\item Introducing Vector LoRA, an efficient adaptation technique of foundation models that addresses both feature hierarchy and gradient flow dynamics.

\item Designing a multi-scale SSIM based reprojection loss function that significantly enhances the performance of our depth estimation approach.

\item Demonstrating superior performance over SOTA methods, showcasing potential and efficacy for depth estimation in surgical contexts.

\end{enumerate}

\section{Methodology}\label{sec:method}
\begin{figure}[!t]%
\centering
\includegraphics[width=1\textwidth]{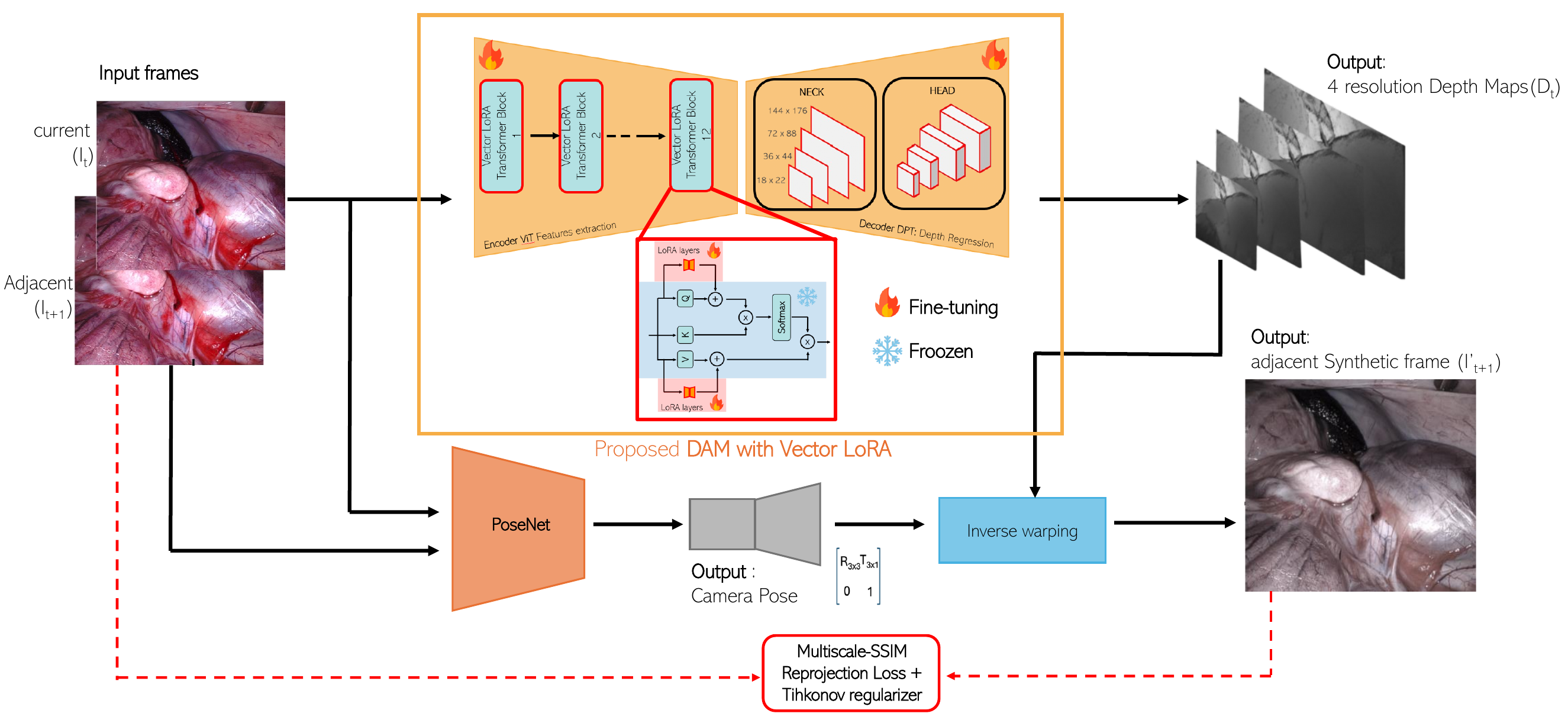}
\caption{Overview of the proposed DARES framework (green box) of DAM v2 with Vector-LoRA (yellow box) for depth estimation and PoseNet for pose estimation which is trained in a self-supervised manner with a multi-scale SSIM based reprojection loss.}
\label{pipeline_schema}
\end{figure}

\subsection{Preliminaries}\label{sec:preliminaries}
\subsubsection{Depth Anything Model V2 (DAM V2)}\label{sec:DAM}
DAM V2 \cite{yang2024depthv2}, features a transformer-based DINOv2 \cite{oquab2023dinov2} encoder for feature extraction and a dense prediction transformer (DPT) \cite{ranftl2021vision} decoder for depth regression. The encoder consists of 12 multi-headed self-attention blocks with alternating multi-layer perceptron (MLP) blocks and normalisation layers. The decoder is composed of two main blocks of neck and head through a series of convolution and upsample layers.

\subsubsection{Low-Rank Adaptation (LoRA)}\label{sec:lora} 

LoRA \cite{hu2022lora} has been introduced to reduce the number of learnable parameters by freezing the original pre-trained weights and adding two learnable low-rank matrices, $A$ and $B$. Drawing on the concept of low "intrinsic rank", LoRA reduces the number of learnable parameters, preventing catastrophic forgetting. 
For a pre-trained weight matrix $W_0 \in \mathbb{R}^{d \times k}$, LoRA constrains its update with a low-rank decomposition: 
$W_0 + \Delta W = W_0 + BA$, where $B \in \mathbb{R}^{d \times r}$ and $A \in \mathbb{R}^{r \times k}$, with the rank $r \ll \min(d, k)$. During training, only $A$ and $B$ are trainable parameters and the rest are frozen. For $h = W_0x$, the modified forward pass using LoRA can be expressed as:
\begin{equation}
\label{eq:lora}
 h = W_0x + \Delta Wx = W_0x + BAx 
\end{equation}
where both $W_0$ and $\Delta W = BA$ are multiplied by the same input $x$, and their output vectors are summed.
These matrices are constructed with dense layers and can be integrated into the query (q), value (v), keys (k), and output (o) vectors in a transformer block (e.g., ViT \cite{dosovitskiy2020vit}). 

\subsubsection{Self-Supervised Depth Estimation}\label{sec:ssl}
One of the most commonly used SSL depth estimation techniques is the reprojection loss approach, as used in the recent SOTA AF-SfMLearner framework  \cite{shao2022self} introduced in the RAS domain. Using the estimated ego-motion between frames and known camera intrinsics, a synthetic image of the temporally adjacent frame is generated. Given the two frames, \( I_t \) and \( I_{t+1} \), the view synthesis can be expressed as $p_{t \rightarrow t+1} = K T_{t+1 \rightarrow t} D_{t+1}(p) K^{-1} p_{t+1}$, where \( p_{t \rightarrow t+1} \) and \( p_{t+1} \) denote the homogeneous coordinates in the source view \( t \) and the target view \( t+1 \) respectively, \( p \) denotes the 2D coordinates, \( K \) denotes the camera intrinsic matrix, \( T_{t+1 \rightarrow t} \) denotes the ego-motion from the target view \( t+1 \) to the source view \( t \), and \( D_{t+1}(p) \) denotes the depth map of target frame \( I_{t+1}(p) \). Then obtain the synthesised frame \( I_{t \rightarrow t+1}(p) \) from the source view \( t \)  as $I_{t \rightarrow t+1}(p) = I_t \left<p_{t \rightarrow t+1} \right>$ where $<.>$ is the bilinear sampling operation as in \cite{shao2022self}. The dissimilarity between the original and synthetic image, known as the reprojection loss $L_{reproj}$, is minimised during training. To address the interframe brightness inconsistency in endoscopic videos, \cite{shao2022self} introduces a regularisation term, the Tihkonov regulariser $L_{reg}$, which operates based on optical flow and appearance flow\cite{shao2022self} calibration. Overall, the regularisation $L_{reg} = L_{rs} + L_{ax} + L_{es}$, where $L_{rs}$,  $L_{ax}$, and $L_{es}$ representing residual-based smoothness loss, auxiliary loss and edge-aware smoothness loss used in \cite{shao2022self}. The total loss, $L_{ssl} = L_{reproj} + L_{reg}$ is a sum of the reprojection loss, $L_{reproj}$ and the regularisation loss,  $L_{reg}$.

\subsection{Proposed Method: DARES}\label{sec:proposed_method}
We propose DARES (Fig. \ref{pipeline_schema}), which consists of DAM with Vector-LoRA and Multi-scale SSIM reprojection loss. The Details of this approach are below.
\subsubsection{Vector-LoRA} 
LoRA helps fine-tune models for specific tasks but fails to account for the fact that earlier layers need more parameters than later ones due to their role in learning general features. To enhance LoRA's effectiveness, we introduce Vector-LoRA, which adaptively adjusts the rank $r$ across the network layers. Higher ranks are assigned to early layers, decreasing progressively through the DAM encoder’s transformer blocks. This strategy comes from the hierarchical nature of neural networks, where initial layers capture generic features and subsequent layers refine these into task-specific details. Therefore, higher initial ranks improve feature adaptation, optimizing resource use. Consequently, our Vector-LoRA can be expressed as:
\begin{equation}
\label{eq:veclora}
 h = W_0x + B_{r}A_{r}x 
\end{equation}
where the dimensions of $B_{r}$, $A_{r}$ are defined by the entries of the rank vector $r$.

\subsubsection{DAM with Vector-LoRA}\label{sec:dam-lora}

In the proposed framework (Fig. \ref{pipeline_schema}), DARES, the encoder architecture of the DAM V2 network, is modified by adding the Vector-LoRA layers inside each of the 12 attention blocks in parallel to the q and v output of the transformer block. In this case, the rank of the Vector-LoRA is:
\begin{equation}
\label{eq:vector}
 r_{vector} = [r_1, r_2, r_3, r_4, r_5, r_6, r_7, r_8, r_9, r_{10}, r_{11}, r_{12}] 
\end{equation}

Where $r_1$ to $r_{12}$ are the rank values/factors that control the number of learnable parameters in each transformer block of the encoder.

\subsubsection{Multiscale-SSIM Reprojection Loss}\label{sec:proposed_loss}

Most works in the field of self-supervised monocular depth estimation such as \cite{godard2019digging,shao2021self} 
utilise a common combination of the $SSIM$ and $L1$ loss to formulate the reprojection error. The $SSIM$ loss,  $S(x,y)$, computes the similarity between images $x$ and $y$ as a function of luminance, contrast, and structure and can be formulated as $S(x, y) = f\left(l(x, y), c(x, y), s(x, y)\right)$ where $l(x,y)$, $c(x,y)$, and $s(x,y)$ refer to the luminance, contrast, and structure similarity respectively\cite{ssim}. To address the particular characteristics of RAS scenes such as highly intricate tissue texture, varying lighting conditions, and motion blur, a more robust method of measuring image similarity is required. In this work, we have utilised multi-scale SSIM \cite{ms-ssim}, that 
processes image pairs by iteratively applying a filter and downsampling them by a factor of 2. At each scale \(j\), the system computes the contrast comparison \(c_j(x, y)\) and the structure comparison \(s_j(x, y)\). The luminance comparison \(l_M(x, y)\) is specifically calculated at the final scale, $M$ \cite{ms-ssim}. The multi-scale SSIM can be written as below:

\begin{equation}
\text{MS-SSIM}(x, y) = [l_M(x, y)]^{\alpha_M} \cdot \prod_{j=1}^{M} [c_j(x, y)]^{\beta_j} [s_j(x, y)]^{\gamma_j}
\end{equation}

$\alpha_M$, $\beta_j$, and $\gamma_j$ are weights used to adjust the relative importance of different components and are taken from \cite{ms-ssim}. The refined reprojection loss utilizing multi-scale SSIM, $L_{\text{ms-reproj}}$, can be found in equation \ref{eq:reproj_hat}. $\alpha$ and $ \beta$ are corresponding weights for each term and were chosen as 0.9 and 0.1 after tuning. 

\begin{equation}
\label{eq:reproj_hat}
L_{\text{ms-reproj}} = \alpha \cdot (1 - MS\text{-}SSIM(I_{\text{target}}, I_{\text{estimate}})) + \beta \cdot |I_{\text{target}} - I_{\text{estimate}}|
\end{equation}

\subsubsection{Self-supervised LoRA Optimisation}\label{sec:optimization}
During training, the pose estimation module, PoseNet \cite{godard2019digging}, takes the current and adjacent frame as input and computes the ego-motion between the two. Meanwhile, the depth prediction module, DAM Vector-LoRA, operates only on the current frame to produce its corresponding 4-resolution depth map. The predicted depth map and camera pose generate a synthetic image of the adjacent frame via reprojection and inverse warping, as explained in section \ref{sec:ssl}. During evaluation, the trained DAM Vector-LoRA and PoseNet estimate depth maps and camera poses, while only the highest resolution depth map is used. 

\section{Implementation Details} \label{sec:impl_details}
\noindent \textbf{Training} Following AF-SfMLearner, we employed the Adam optimiser with a step decay learning rate, starting at 0.0001 and decaying every 10 steps, for a total of 50 epochs. We used a batch size of 12 and trained the model on an A6000 GPU, completing the process in 20 hours. The proposed pipeline was developed in PyTorch. After tuning, the rank vector $r$ was chosen as $r=[14,14,12,12,10,10,8,8,8,8,8,8]$.

\noindent \textbf{Dataset} We validate our model using the SCARED dataset\cite{allan2021stereo}, which comprises 35 endoscopic sequences derived from porcine cadavers. Following the Eigen-Zhou evaluation protocol established in \cite{Eigen2014,Zhou2017}, 15351 frames were used for training, 1705 for validation, and 551 for testing. For a fair comparison, ego-motion was evaluated using two consecutive trajectories of length 410 and 833 frames defined in \cite{shao2022self}. To ensure consistency and manageability, all images were resized to 320x256 pixels. 

\noindent \textbf{Evaluation Protocol} The performance of our pipeline is evaluated against several methods, including DeFeat-Net \cite{defeat}, SC-SfMLearner \cite{scsfm}, Monodepth2 \cite{godard2019digging}, Endo-SfM \cite{endoslam} and AF-SfMLearner \cite{shao2021self}. The quantitative results of these baseline methods is obtained from \cite{shao2021self}. The depth estimation module is evaluated in terms of absolute relative error (Abs Rel), squared relative error (Sq Rel), Root Mean Square Error (RMSE), Root Mean Square Error in the logarithmic space (RMSE log), and Threshold Accuracy ($\delta$), while to evaluate the pose estimation module the Absolute Trajectory Error (ATE) is used, following \cite{shao2021self}. We adopt the reported metrics for all the comparing methods except for the most recent baseline, AF-SfMLearner \cite{shao2021self}, where we reproduce the results in our environment setting.

\begin{table}[!b]
    \centering
    \caption{Comparison of benchmark methods on depth estimation and pose estimation matrices}
     \resizebox{\textwidth}{!}{
    \begin{tabular}{lccccccc}
        \toprule
        \textbf{Method} & \textbf{Abs Rel $\downarrow$} & \textbf{sq Rel $\downarrow$} & \textbf{RMSE $\downarrow$} & \textbf{RMSE log $\downarrow$} & \textbf{$\delta \uparrow$} & \textbf{ATE-Traj1 $\downarrow$} & \textbf{ATE-Traj2 $\downarrow$} \\
        \midrule
        DeFeat-Net \cite{defeat} & 0.077 & 0.792 & 6.688 & 0.108 & 0.941 & 0.1765 & 0.0995  \\
        SC-SfMLearner \cite{bian2019unsupervised} & 0.068 & 0.645 & 5.988 & 0.097 & 0.957 & 0.0767 & 0.0509  \\
        Monodepth2 \cite{godard2019digging} & 0.071 & 0.590 & 5.606 & 0.094 & 0.953 & 0.0769 & 0.0554 \\
        Endo-SfM \cite{endoslam} & 0.062 & 0.606 & 5.726 & 0.093 & 0.957 & 0.0759 & 0.0500  \\
        AF-SfMLearner \cite{shao2022self} & 0.060 & 0.477 & 5.100 & 0.083 & 0.966 & 0.0757 & 0.0501  \\
        Zero-Shot DAM V2 & 0.091 & 1.056 & 7.601 & 0.126 & 0.916 &  -  & - \\
        Fully fine-tuned DAM V2 & 0.076 & 0.742 & 6.344 & 0.108 & 0.937 & \textbf{0.0749} & 0.0510\\
        \textbf{Ours (DARES)} & \textbf{0.052} & \textbf{0.356} & \textbf{4.483} & \textbf{0.073} & \textbf{0.980} &  0.0752 & \textbf{0.0498} \\
        \bottomrule
    \end{tabular}}
    \label{res_tab}
\end{table}

\section{Results}\label{sec:results}

\begin{figure}[!t]
\centering
\includegraphics[width=0.9\textwidth]{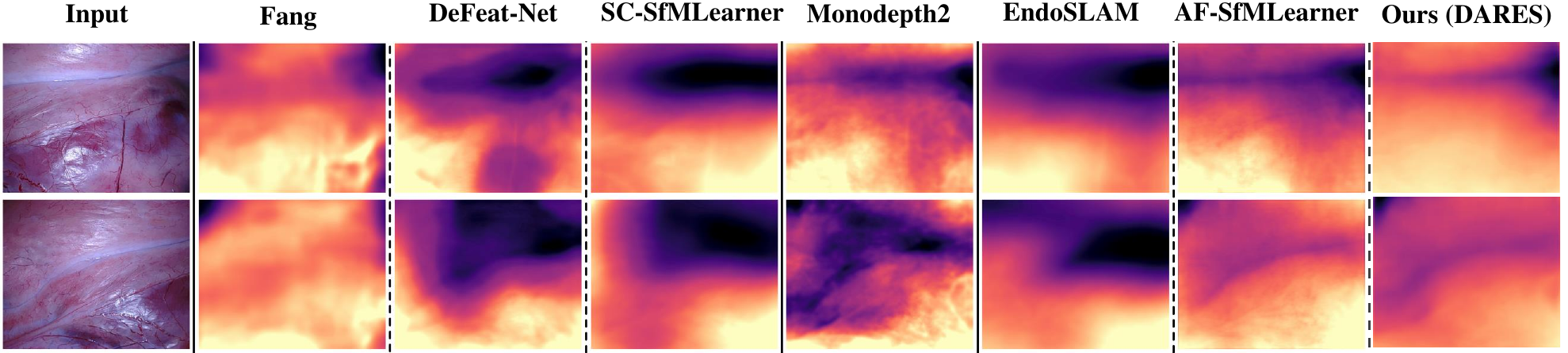}
\caption{Qualitative comparison of our depth estimation with SOTA baselines}
\label{comparison_depth}
\end{figure}

We evaluate the depth estimation accuracy of our framework against several SOTA methods. Table \ref{res_tab} shows the quantitative results, demonstrating that our pipeline outperforms all other SOTA methods in the depth estimation task, achieving an improvement of 13.3\% over the second-best approach.  
The results of the Zero-Shot DAM V2 and Fully fine-tuned DAM V2 models are especially notable. Due to the large domain shift between the data that the foundation model DAM V2 has originally been trained on, and our target endoscopy scenes, Zero-Shot DAM V2 fails to generalise to this distinct environment and is unable to improve upon the SOTA. On the other hand, simply fully fine-tuning this model, results in suboptimal performance caused by catastrophic forgetting\cite{catastrophic}, where the model overfits to the new data and skews the learned parameters, hindering its robustness and generalisation. These observations highlight the necessity of a strategic approach to adapt this foundation model to our specific case, as in our pipeline, since simply using or fine-tuning a foundation model will not unlock its full potential.

\begin{figure}[!t]%
\centering
\includegraphics[width=0.9\textwidth]{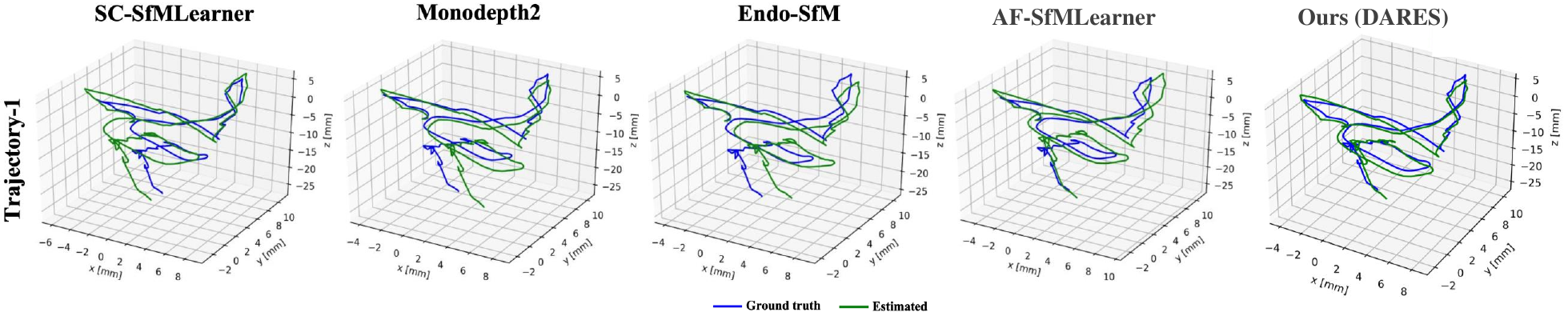}
\caption{Comparison of ego-motion predicted by our model with SOTA baselines}
\label{comparison_pose}
\end{figure}

\begin{table}[!b]
    \begin{minipage}{0.42\linewidth}
        \centering
        \includegraphics[width=1\textwidth]{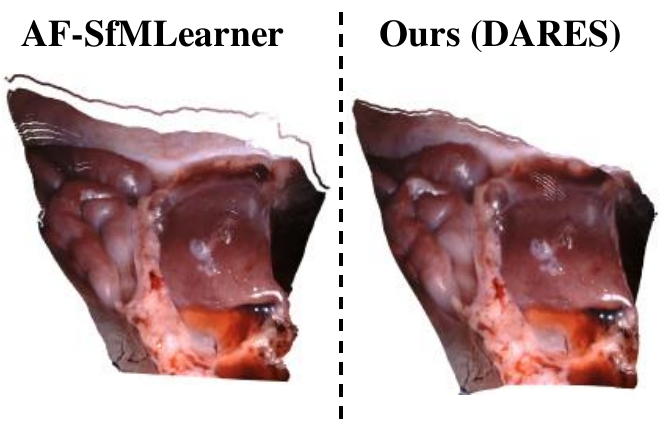}
        \captionof{figure}{Qualitative 3D reconstruction comparison of our model with AF-SfMLearner.}
        \label{3drec}
    \end{minipage}\hfill
    \begin{minipage}{0.55\linewidth}
        \centering
        \caption{Time efficiency of our method vs. AF-SfMLearner.}
        \scalebox{0.73}{
        \begin{tabular}{lccc}
            \toprule
            \textbf{Method} & \textbf{Total.(M) $\downarrow$} & \textbf{Train.(M) $\downarrow$} & \textbf{Speed(ms) $\downarrow$} \\
            \midrule
            AF-SfMLearner \cite{shao2022self} & 14.8 & 14.8 & 8.0 \\
            \textbf{Ours (DARES)} & 24.9 & 2.88 & 15.6 \\
            \bottomrule
        \end{tabular}}
        \label{speed_tab}
    \end{minipage}
\end{table}

A qualitative evaluation of our depth estimation results is pictured in Fig. \ref{comparison_depth}. It can be observed that DARES is better at capturing finer depth declines. In the pose estimation task, given the same pose network for both AF-SfMlearner \cite{shao2022self} and fully fine-tuned DAM V2, we get comparable results, for Trajectory 1, and slightly better results for Trajectory 2 (Table \ref{res_tab}).

Fig. \ref{comparison_pose} displays the trajectories predicted by different benchmarks for Trajectory-1 against our framework. Compared to the most recent SOTA, AF-SfMLearner, our pipeline performs better in estimating the ego-motion of the endoscopic camera. Finally, to demonstrate the 3D reconstruction capabilities of our model, Fig. \ref{3drec} presents a sample 3D scene reconstructed using our approach compared to AF-SfMLearner. DARES exhibits fewer artifacts than the SOTA technique, resulting in a more stable reconstruction.

\begin{table}[!t]
    \centering
    \caption{Effects of utilizing DAM V2,  Vector LoRA (V-LoRA), and MS-SSIM based reprojection loss.}
    \scalebox{0.8}{
    \begin{tabular}{c c c c c c c c}
        \toprule
        \textbf{DAM} & \textbf{LoRA} & \textbf{V-LoRA} & \textbf{MS-SSIM} & \textbf{Abs Rel $\downarrow$} & \textbf{sq Rel $\downarrow$} & \textbf{RMSE $\downarrow$} & \textbf{RMSE log $\downarrow$} \\
        \midrule
        $\times$ & $\times$ & $\times$ & $\times$ & 0.060 & 0.477 & 5.100 & 0.083 \\
         $\checkmark$ & $\times$ & $\times$ & $\times$ &0.076 & 0.742 & 6.344 & 0.108 \\
         $\checkmark$ & $\checkmark$ & $\times$ & $\checkmark$ &0.053 & 0.372 & 4.565 & 0.074 \\
        $\checkmark$ & $\times$ & $\checkmark$ & $\checkmark$ & 0.052 & 0.356 & 4.483 & 0.073 \\
        \bottomrule
    \end{tabular}}
    \label{abl_tab}
\end{table}

To highlight the impact of each of our pipeline components, a series of ablation experiments have been carried out (Table \ref{abl_tab}). The results show that using DAM V2 in a surgical endoscopy setting coupled with LoRA and our designed reprojection loss will result in 11.6\% improvement in the performance of our self-supervised monocular depth estimation framework compared to SOTA approaches. If this is replaced by utilizing our proposed Vector-LoRA instead of LoRA, an improvement of 13.3\% can be reached, reducing the depth estimation error from 0.076 to 0.052. To demonstrate the time efficiency and size of our model compared to the SOTA AF-SfMLearner\cite{shao2022self} approach,
Table \ref{speed_tab} compares the number of total parameters, i.e. complexity of the network, w.r.t the number of trainable parameters. Our approach has less trainable parameters (12\% of the total), significantly reducing the training time. The inference time of our model is 15.6 ms, which is approximately 64 fps and is close to real-time speed.

\section{Conclusion}\label{sec:conclusion}

We present the DARES framework for monocular self-supervised depth estimation, utilizing the DAM V2 vision foundation model adapted to RAS scenes. We have introduced Vector-LoRA for efficient adaptation of DAM V2 and designed a multi-scale SSIM based reprojection loss for robust depth map and surface reconstruction. Our results and ablation studies have demonstrated the effectiveness of both Vector-LoRA and the multi-scale SSIM based reprojection loss, providing compelling evidence of the successful adaptation of the foundation model to the surgical domain.
Future work will explore methods to enhance the robustness and reliability of foundation models in endoscopic scenes, making them more suitable for surgical applications by leveraging all available RAS datasets and broader adaptation techniques like GaLore \cite{zhao2024galore} and MoRA \cite{jiang2024mora}.

\section*{Acknowledgement}
This work was carried out as part of the UCL Medical Image Computing Summer School (MedICSS) and has been supported in part by the Wellcome/EPSRC Centre for Interventional and Surgical Sciences (WEISS) [203145/Z/16/Z]; Engineering and Physical Sciences Research Council (EPSRC) [EP/W00805X/1, EP/Y01958X/1, EP/P012841/1]; the Department of Science, Innovation and Technology (DSIT); the Royal Academy of Engineering under the Chair in Emerging Technologies programme; Horizon 2020 FET [GA863146]; i4health EP/S021930/1; UKRI Training Grant {EP/S021612/1}; Medical Microinstruments, Inc., Wilmington, USA and the Italian Ministry of Universities and Research (NRRP DM 117); the ANTHEM project, funded by the National Plan for NRRP Complementary Investments (CUP: B53C22006700001); and the Multilayered Urban Sustainability Action (MUSA) project (ECS00000037), funded by the European Union – NextGenerationEU, under the National Recovery and Resilience Plan (NRRP). For the purpose of open access, the authors have applied a CC BY public copyright licence to any author accepted manuscript version arising from this submission.

\bibliography{ref.bib}{}
\bibliographystyle{splncs04}
\end{document}